\documentclass[10pt,twocolumn,letterpaper]{article}

\usepackage{cvpr}              

\usepackage{graphicx} 
\usepackage{xcolor}

\definecolor{cvprblue}{rgb}{0.21,0.49,0.74}
\usepackage[pagebackref,breaklinks,colorlinks,allcolors=cvprblue]{hyperref}

\definecolor{myblue}{rgb}{0.1, 0.2, 0.7}
\definecolor{mylightblue}{rgb}{0.68, 0.85, 0.90}

\newcommand\blfootnote[1]{%
  \begingroup
  \renewcommand\thefootnote{}\footnote{#1}%
  \addtocounter{footnote}{-1}%
  \endgroup
}
\renewcommand{\footnoterule}{%
  \kern -3pt               
  \hrule width \columnwidth height 0.4pt 
  \kern 2.6pt              
}

\def\paperID{9801} 
\def\confName{CVPR}
\def\confYear{2026}

\title{ViSA: 3D-Aware Video Shading for Real-Time Upper-Body Avatar Creation}

\author{
  Fan Yang\textsuperscript{1*} \quad
  Heyuan Li\textsuperscript{2*} \quad
  Peihao Li\textsuperscript{3} \quad
  Weihao Yuan\textsuperscript{3}  \quad 
  Lingteng Qiu\textsuperscript{3} \quad
  Chaoyue Song\textsuperscript{1} \quad  \\[\smallskipamount]
  Cheng Chen\textsuperscript{1} \quad
  Yisheng He\textsuperscript{3} \quad
  Shifeng Zhang\textsuperscript{3$\dag$} \quad
  Xiaoguang Han\textsuperscript{2} \quad
  Steven Hoi\textsuperscript{3} \quad
  Guosheng Lin\textsuperscript{1$\dag$}
  \vspace{0.2cm} \\ 
  {
  \textsuperscript{1}Nanyang Technological University \quad
  \textsuperscript{2}The Chinese University of Hong Kong, Shenzhen 
  } \\[\smallskipamount] 
  {
  \textsuperscript{3} Tongyi Lab, Alibaba Group
  }
}

\begin{document}

\twocolumn[{%
\renewcommand\twocolumn[1][]{#1}%
\maketitle
\begin{center}
\centering
    \vspace{-0.5em}{\includegraphics[width=1.0\linewidth]{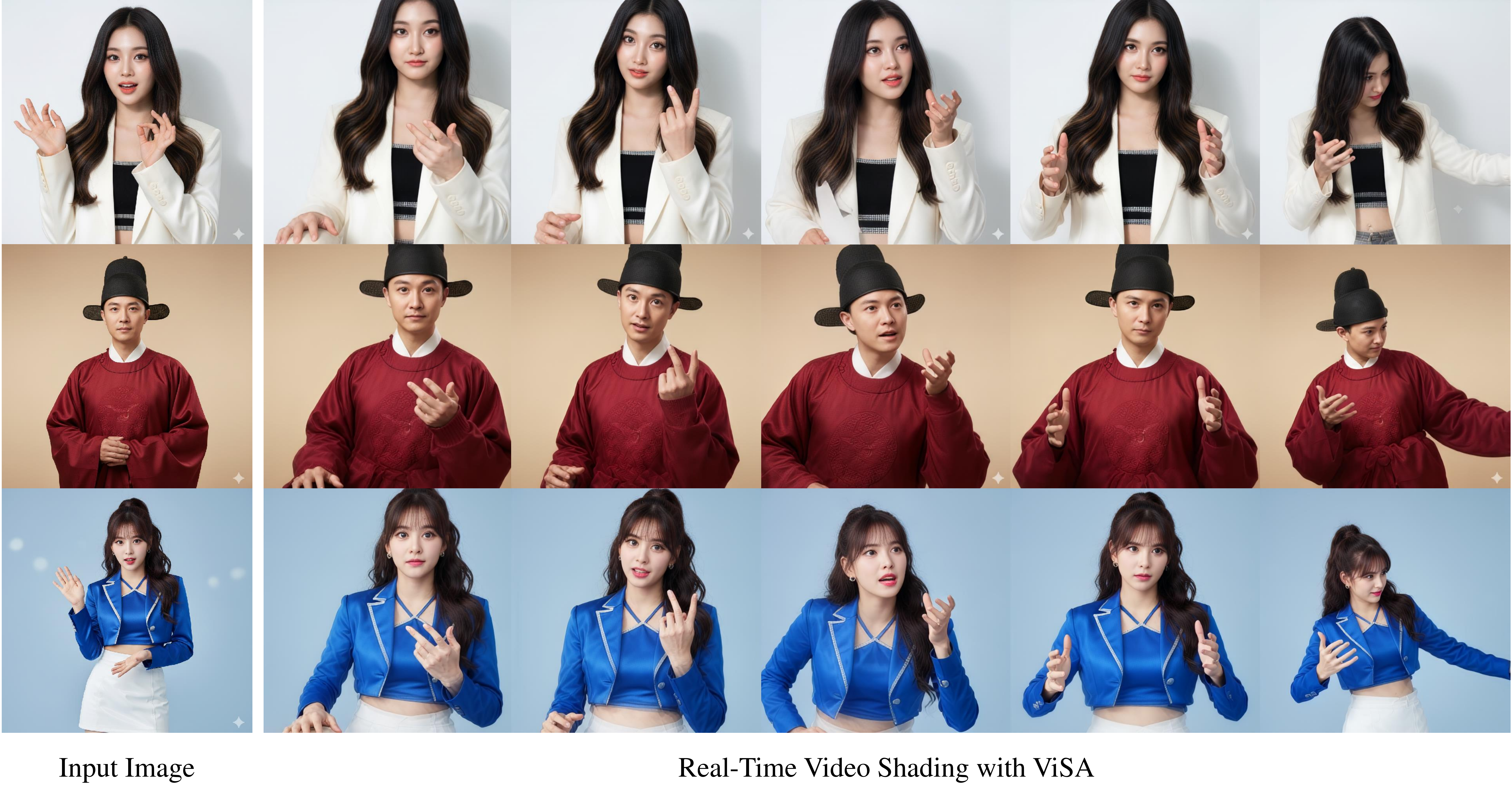}}
    \captionof{figure}{\textbf{Photorealistic, consistent, and controllable character animation from a single reference image}. Our method enables photorealistic upper-body avatar generation that preserves appearance fidelity across diverse poses and expressions while maintaining temporal coherence in real-time video synthesis.
    Input images are synthesized using Gemini~\cite{team2023gemini}.}
    \label{fig:vis}
\end{center}
\vspace{-0.5em}
}]

\blfootnote{%
  \begin{tabular}{@{} l @{\hspace{0.5em}} l} 
    \textsuperscript{*} & Equal contribution, work done during an internship at Tongyi Lab. \\
    \textsuperscript{\textdagger} & Corresponding author.
  \end{tabular}%
}

\newcommand{\fref}[1]{Fig.~\ref{#1}}
\newcommand{\Fref}[1]{Figure~\ref{#1}}
\newcommand{\sref}[1]{Sec.~\ref{#1}}
\newcommand{\Sref}[1]{Section~\ref{#1}}

\begin{abstract}

Generating high-fidelity upper-body 3D avatars from one-shot input image remains a significant challenge. Current 3D avatar generation methods, which rely on large reconstruction models, are fast and capable of producing stable body structures, but they often suffer from artifacts such as blurry textures and stiff, unnatural motion. In contrast, generative video models show promising performance by synthesizing photorealistic and dynamic results, but they frequently struggle with unstable behavior, including body structural errors and identity drift. To address these limitations, we propose a novel approach that combines the strengths of both paradigms. Our framework employs a 3D reconstruction model to provide robust structural and appearance priors, which in turn guides a real-time autoregressive video diffusion model for rendering. This process enables the model to synthesize high-frequency, photorealistic details and fluid dynamics in real time, effectively reducing texture blur and motion stiffness while preventing the structural inconsistencies common in video generation methods. By uniting the geometric stability of 3D reconstruction with the generative capabilities of video models, our method produces high-fidelity digital avatars with realistic appearance and dynamic, temporally coherent motion. Experiments demonstrate that our approach significantly reduces artifacts and achieves substantial improvements in visual quality over leading methods, providing a robust and efficient solution for real-time applications such as gaming and virtual reality. \textcolor{blue!60!white}{https://lhyfst.github.io/visa}.

\end{abstract}    
\section{Introduction}
\label{sec:intro}


The request to generate real-time, expressive digital avatars from one-shot input has seen significant progress. 3D reconstruction methods, leveraging representations like Neural Radiance Fields or 3D Gaussian Splatting~\cite{hong2023lrm,qiu2025LHM,tang2025lgm, he2025lam, GUAVA}, excel at providing a stable body structure. By binding the reconstruction to parametric human templates~\cite{smplx:2019}, they can produce dynamic results. Yet, these methods struggle to achieve true photorealism. The inherent ill-posed nature of single-view reconstruction often leads to artifacts such as blurry textures and incomplete geometry that persist in the rendering pipeline, degrading the final output and making the avatars appear unrealistic. 

On the other hand, video diffusion models~\cite{wan2025, kong2024hunyuanvideo, gao2025seedance} have demonstrated promising capabilities in synthesizing photorealistic videos with rich details and high natural dynamics. 
Existing methods adapt pretrained video diffusion models to generate video avatars with different condition guidance, such as keypoints~\cite{hu2024animate}, DensePose~\cite{guler2018densepose, xu2024magicanimate}, and rendered SMPL meshes~\cite{zhu2024champ}. 
While these guidance enhance structural accuracy, they still suffer from the loss of fine-grained details like texture or clothing, thus struggling to preserve identity consistency. Moreover, the slow inference speed of these methods also makes them impractical for generating long-duration and interactive avatars. 



To address these problems, we explore a way of generating realistic avatars by combining an explicit 3D reconstruction process with an implicit 3D-aware video rendering process, capitalizing on the complementary strengths of both methods. The core motivation is that explicit 3D reconstruction largely preserves the identity, shape, and texture of the reference avatar, providing a high-fidelity prior for the subsequent generative stage, which alleviates generation uncertainty and enables lightweight, few-step autoregressive models for real-time controlled video synthesis. Specifically, 
our method first extracts multi-level 2D features from the input avatar image. These features are then lifted into a 3D feature space via a lightweight reconstruction network, animated and projected back onto the image plane to guide the rendering process. To achieve real-time and long-sequence rendering, we propose to build the implicit rendering process with a pre-distilled, few-step causal video diffusion model~\cite{huang2025self}. Specifically, we repurpose it as a powerful neural shader by finetuning it with 3D-aware features rendered from our reconstruction model. This is non-trivial, as naively finetuning the distilled model with a standard denoising objective leads to blurry outputs, which is caused by the high variance of the optimization target~\cite{luo2025adding, luo2025noise}. To prevent the model from deviating from its learned distribution, we propose an adversarial distribution preservation loss that enhances generation fidelity, leading to significantly sharper outputs with realistic high-frequency details.

To train our model, we curate a dataset from Speaker-5M~\cite{zhang2025speakervid}, a large-scale dataset of human speaking videos. We process this data by annotating it with SMPL-X and FLAME parameters to support our 3D-aware framework. 
Extensive experiments demonstrate that our method significantly outperforms existing approaches, generating photorealistic and temporally coherent animation results from a single image in real-time. Our contributions can be summarized as follows:
i) We propose a novel framework that integrates one-shot 3D avatar reconstruction with a real-time autoregressive video shader, unifying geometric stability with generative rendering power.
ii) We introduce a novel adversarial distribution preservation loss, which enables the efficient finetuning of a pre-distilled, few-step video diffusion model to act as a powerful neural renderer while mitigating the typical quality degradation.
iii) Our method establishes a new state-of-the-art in real-time upper-body avatar generation, achieving superior visual quality and motion realism, validated by comprehensive qualitative and quantitative comparisons.

\section{Related Work}
\label{sec:related_work}

\begin{figure*}
  \centering
  \includegraphics[width=\linewidth]{sec/imgs/fig2_pipeline_v8_0.pdf}

  \caption{\textbf{Overview of the proposed ViSA}. In the first stage, we train a one-shot, feed-forward transformer to regress a 3D Gaussian avatar in canonical space, conditioned on geometric, semantic, and low-level embeddings. In the second stage, we employ an autoregressive video model as a video renderer, conditioned on the 3D-aware features from stage one, to generate photorealistic results in real time.} 
   \label{fig:pipeline}
\end{figure*}

\subsection{One-shot 3D Human Reconstruction} 
Reconstructing animatable avatars from one-shot input remains a difficult task. Early approaches typically relied on parametric body models~\cite{alldieck2018videobasedreconstruction3d}. 
Recently, feedforward methods~\cite{qiu2025LHM,zhuang2024idol,qiu2025pf, he2025lam,qiu2024AniGS} use scalable transformer architectures to directly regress human avatars in a predefined canonical space, typically conditioned on a parametric head model FLAME~\cite{FLAME:SiggraphAsia2017} or full body model SMPL-X~\cite{smplx:2019}. 
GUAVA~\cite{GUAVA} leverages inverse texture mapping and projection sampling to generate animatable upper-body avatars with an expressive human parametric model. Although these approaches have advanced human reconstruction, the scarcity of high-quality 3D large-scale data supervision or precisely aligned training annotations often leads to blurred outputs and artifacts in fine-grained regions, thereby degrading the overall realism and perceptual quality.

\subsection{Controllable Human-Centric Video Generation} 
Fueled by large scale datasets and advances in diffusion architectures~\cite{rombach2022high, peebles2023scalable}, diffusion-based human-centric video generation approaches~\cite{wan2025, kong2024hunyuanvideo, gao2025seedance, team2025longcat, ali2025world} have progressed substantially, especially in generalization and high-fidelity visual quality. 
Animate Anyone~\cite{hu2024animate} and EMO~\cite{tian2024emo} utilizes reference network to preserve the subject's identity and employs a lightweight pose-conditioned module for guidance. Subsequent works, such as Champ~\cite{zhu2024champ}, MIMO~\cite{men2024mimo}, and Human4DiT~\cite{shao2024human4dit}, replace sparse 2D keypoints with 3D shape guidance~\cite{smplx:2019,loper2015smpl}, enabling more accurate and controllable generation. Realisdance~\cite{zhou2024realisdance} utilizes a variety of control signals, such as sparse keypoints and rendered image from MANO~\cite{romero2022embodied} and SMPL~\cite{loper2015smpl} model,  to achieve more controllable generation effects. Although they can produce vivid human animation videos, they struggle to generate long-consistent videos with stable body structures.  
Additionally, these methods require several minutes to generate a video sequence given the human poses \cite{zhu2024champ,shao2024human4dit}, limiting their practicality in interactive applications.

\subsection{Autoregressive Video Diffusion Model}  
Autoregressive diffusion model~\cite{chen2024diffusion, jin2024pyramidal, chen2025skyreels, gu2025long}, which generates video frames sequentially, is a promising avenue to reduce the latency of video generation. Recent advances such as CausVid~\cite{yin2025slow} and Self-forcing~\cite{huang2025self} demonstrate real-time high-fidelity video generation capabilities by integrating the distillation of few step diffusion models with the adaption of causal inference. Our model is also built on autoregressive video diffusion model, but reformulatess it as a real-time video shader. By conditioning the autoregressive video diffusion model with dense 3D aware features, our methods achieves photorealistic rendering in real time.

\section{Method}
\label{sec:method}


\subsection{Overview}
\label{sec:overview}

We present ViSA, a novel framework for generating real-time, photorealistic avatars from one-shot image. Our approach is designed as a two-stage pipeline that integrates the geometric stability of 3D reconstruction with the generative power of a real-time autoregressive video diffusion model.
In the following sections, we will detail the architecture of our 3D reconstruction model in \sref{sec:3d_recon}, the finetuning of the autoregressive video renderer in \sref{sec:video_render}, the comprehensive training strategy in \sref{sec:train_strategy}, and the dataset curation process in \sref{sec:dataset}.

\subsection{3D Reconstruction}
\label{sec:3d_recon}

To construct a comprehensive guide for the reconstruction and rendering, we extract and fuse three complementary types of features:
i) \textit{Semantic Features ($F_{sem}$).}
We employ a pretrained DINOv2 encoder~\cite{oquab2023dinov2}, denoted as $E_{dino}$, to capture high-level semantic information about the subject. 
The semantic feature is extracted as
\begin{equation}
F_{sem} = E_{dino}(I_{dino}),
\end{equation}
where $I_{dino}$ is the input image $I$ resized to fit the $E_{dino}$.

\noindent ii) \textit{Low-Level Visual Features ($F_{vis}$).} DINOv2 is not optimized for reconstruction, as it tends to discard fine-grained local details and is constrained to relatively low input resolutions. To address this problem, we involve the pretrained VAE encoder from a video diffusion model~\cite{wan2025}, denoted as $E_{vae}$, to provide complementary information. As this encoder is trained for reconstruction, it excels at preserving high-fidelity local information and can process much higher-resolution inputs. We extract multi-scale feature maps $\{f_{vae}^l\}_{l \in L}$ from a set of intermediate layers $L$ of the encoder. These features are then aggregated by a pyramid encoding network, $Net_{pyr}$, to produce a fused visual feature map:
\begin{equation}
F_{vis} = Net_{pyr}(\{f_{vae}^l\}_{l \in L}).
\end{equation}

\noindent iii) \textit{General Human Prior ($F_{prior}$).} While the features above provide person-specific information, we also introduce a person-invariant prior to encode general human structure. To achieve this, we create a learnable embedding table for the vertices of the SMPL-X template. 
For each of the $N_v$ vertices in the canonical SMPL-X model, we assign a learnable $d$-dimensional feature vector. This collection of embeddings forms a human prior feature table, which is independent of the input image and shared across all subjects:
\begin{equation}
F_{prior} = \{e_i\}_{i=1}^{N_v}, \quad \text{where } e_i \in \mathbb{R}^{d}.
\end{equation}




We merge these multi-modal features in 3D space to predict the attributes of the 3D Gaussian Splatting (GS) representation as well as the 3D aware features to guide the following generation process. 
Specifically, we first lift the 2D features, $F_{sem}$ and $F_{vis}$, into the 3D canonical space of the SMPL-X model with the estimated pose and shape parameters of the input image. 
Denoting these lifted features for the $i$-th vertex as $f_{sem, i}$ and $f_{vis, i}$, they are then concatenated with the per-vertex learnable prior $e_i$ to formulate the final feature token $T_i$:
\begin{equation}
T_i = \text{concat}(f_{sem, i}, f_{vis, i}, e_i).
\end{equation}

\noindent The set of tokens $\{T_i\}_{i=1}^{N_v}$ are then processed by a light-weight reconstruction network, which consists of five transformer layers, to predict the 3D Gaussian attributes and 3D-aware features. 

The overall structure of our reconstruction model is illustrated in Fig.~\ref{fig:pipeline}. 
During training, our model is supervised by a composite loss function, including rgb and mask supervision, which could be formulated as:
\begin{equation}
\begin{split}
\mathcal{L}_{recon} = \lambda_{L1} \|I_{ren} - I_{gt}\|_1 &+ \lambda_{p} \mathcal{L}_{LPIPS}(I_{ren}, I_{gt}) \\
& + \lambda_{mask} \|M_{ren} - M_{gt}\|_2^2,
\end{split}
\end{equation}
where $\lambda_{L1}$, $\lambda_{p}$, and $\lambda_{mask}$ are weighting coefficients. This 3D reconstruction model provides the following video shader with 3D-aware guidance, which is essential for stable avatar generation.

\subsection{Real-Time Video Shader}
\label{sec:video_render}

We build our video shader upon Self-forcing~\cite{huang2025self}, a state-of-the-art few-step autoregressive video diffusion model for high-quality, real-time video synthesis. We introduce several key architectural modifications to adapt it into a powerful, 3D-aware neural renderer.

\subsubsection{Dynamic Conditioning with 3D-Aware Features}
\label{sec:dc}
Traditional controllable generation methods~\cite{hu2024animate, xu2024magicanimate, zhu2024champ, zhou2024realisdance} typically rely on 2D guidance signals, such as keypoints or rendered SMPL(-X) images. These RGB-based conditions are encoded into the latent space using a pretrained VAE encoder. We argue that this paradigm has two fundamental limitations:
\textbf{Information Bottleneck}: 2D guidance is either sparse (keypoints) or semantically simplistic (SMPL). They primarily convey motion, offering weak priors for subject-specific shape and texture, which often leads to identity drift.
\textbf{Latency Bottleneck}: In real-time settings, the necessity of a forward pass through the VAE encoder for each conditioning frame introduces significant latency, hindering performance.
In contrast, we propose conditioning the video model directly on the 3D-aware feature maps from our reconstruction model. 
These dense features encode rich, multi-level information about identity, shape, and texture, providing a strong prior that mitigates identity drift.
We integrate this guidance via channel-wise concatenation, which directly concatenates the 3D-aware features with noisy latents before feeding them into the diffusion model, which is an efficient strategy requiring minimal extra parameters.
Our approach not only supplies rich semantic priors for generation but also bypasses VAE encoding, yielding a more informative and efficient solution.

\subsubsection{Static Conditioning with a Reference Image}
To further anchor the avatar's identity, we condition the generation on the original input image. The Self-forcing model, which is distilled from WAN2.1 1.3B~\cite{wan2025}, is designed for text-to-video generation and lacks a native mechanism for image conditioning. However, its causal autoregressive structure offers an elegant solution. Specifically, we treat the VAE-encoded reference image as the first frame of the generation sequence. We perform a single forward pass to compute its Key-Value (KV) cache within the model's attention layers. This KV cache is then held constant and reused throughout the subsequent autoregressive generation process, serving as a persistent identity reference. To address the spatial misalignment between the static reference frame and the animated subsequent frames, we adopt the shifted Rotary Position Embedding (RoPE)~\cite{su2024roformer, zhou2024realisdance}, which ensures coherent feature integration despite significant motion, effectively injecting the reference identity into the entire generated video.

\subsubsection{Finetuning Strategy}
Our finetuning process is designed to effectively adapt the pre-distilled video diffusion model into a high-fidelity, 3D-aware video renderer. 
This is achieved through a combination of a paired condition supervision loss, a novel adversarial loss to maintain sharpness, and a progressive training strategy.

\noindent\textbf{Paired Condition Supervision.}
Given a training triplet $(I_{ref}, V_{gt}, \Theta_{gt})$, which includes a reference image, ground-truth video frames, and the corresponding SMPL-X parameters, our pipeline proceeds as follows: First, the 3D reconstruction module generates a canonical Gaussian avatar from $I_{ref}$. This avatar is then animated using the target poses from $\Theta_{gt}$. For each frame, we render a 3D-aware feature map $F_{cond}$ to serve as the dynamic condition and adopt $V_{gt}$ to provide paired supervision. During finetuning, we randomly sample a timestep from the predefined few-step set of Self-forcing and then generate video frames $\hat{V}$ with a self-rollout training strategy.

To guide the generation, we apply supervision in both the latent and pixel spaces. For latent-space supervision, we adopt MSE loss to enforce content consistency between generated frames and groundtruth frames. For pixel-space supervision, which ensures the preservation of fine-grained details, we use a combination of pixel-wise, perceptual, and style losses. To mitigate the computational cost of decoding for this pixel-level comparison, we employ a tiny decoder~\cite{BoerBohan2025TAEHV}, which is distilled from the WAN VAE~\cite{wan2025} with minimal performance degradation,  as our VAE decoder. The final paired condition guidance loss $\mathcal{L}_{\text{pair}}$ is formulated as a weighted sum of these components:
\begin{equation}
\label{eq:paired_loss}
\begin{split}
\mathcal{L}_{\text{pair}} ={}& \lambda_{\text{latent}} \|Z_{\hat{V}} - Z_{V_{\text{gt}}}\|_2^2 + \lambda_{L2} \|\hat{V} - V_{\text{gt}}\|_2^2 \\
&+ \lambda_{p} \mathcal{L}_{\text{LPIPS}}(\hat{V}, V_{\text{gt}}) + \lambda_{s} \mathcal{L}_{\text{style}}(\hat{V}, V_{\text{gt}}),
\end{split}
\end{equation}
where $\lambda_{\text{latent}}$, $\lambda_{L2}$, $\lambda_{p}$, and $\lambda_{s}$ are loss weights.

\noindent\textbf{Adversarial Distribution Preservation Loss.}
During our experiments, we found naively finetuning the distilled video diffusion model with the denoising objective ($\mathcal{L}_{pair}$), while capable of injecting condition guidance, results in blurry outputs. This is caused by the high variance between randomly sampled noise and fixed ground-truth. As a result, the model tend to predict an average of multiple plausible images~\cite{luo2025adding,luo2025noise}. To counteract this and enforce the sharpness and realism of the generation results, we introduce an adversarial training scheme to preserve the finetuned model from falling out the distribution of real-world dataset.

\begin{figure}[t]
  \centering
  \includegraphics[width=0.45\textwidth]{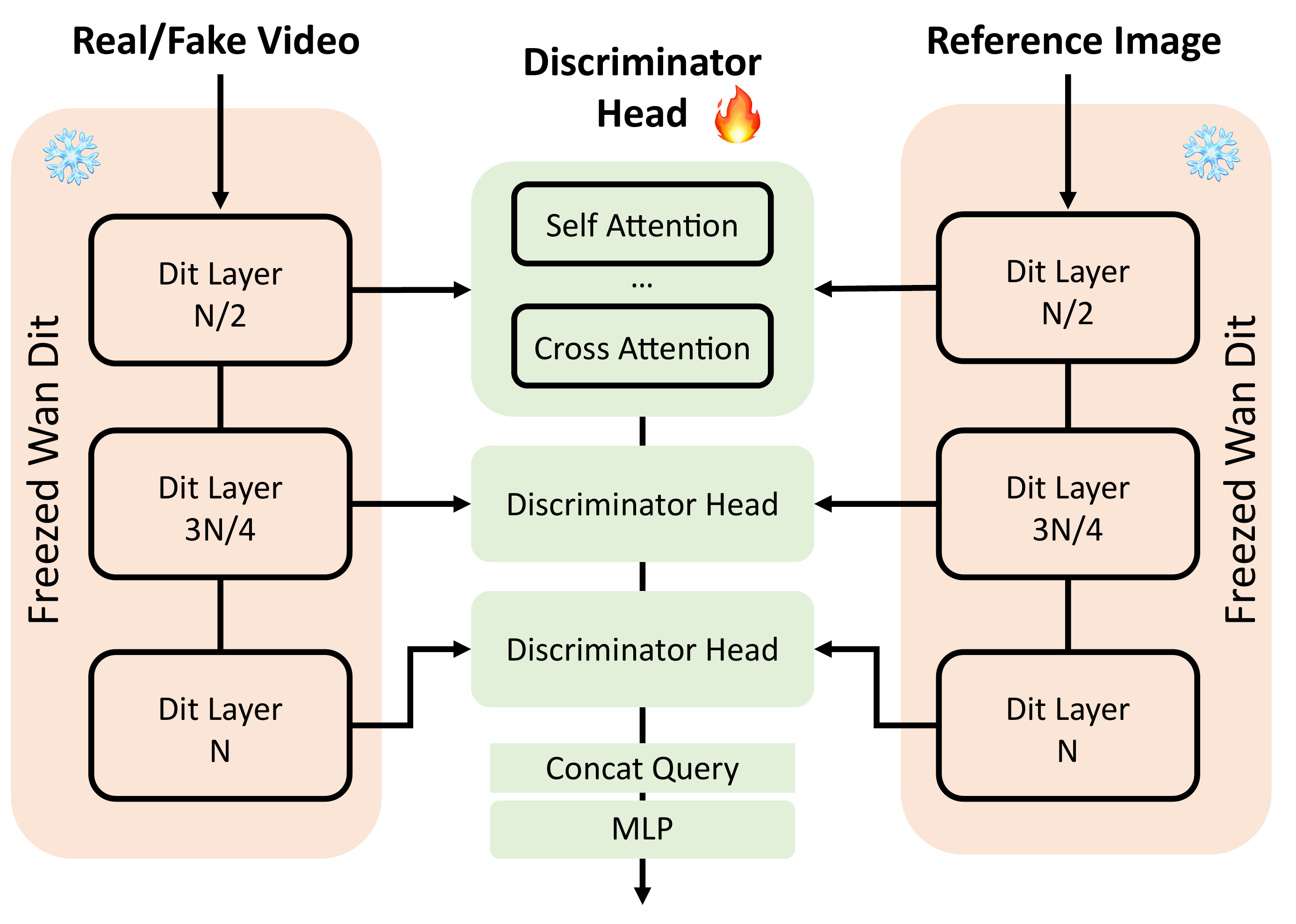}

  \caption{\textbf{The architecture of our discriminator}. We adopt pretrained WAN video diffusion model as our backbone and add trainable classification head on it to predict logit. We also involve the feature from the reference image to enhancing the training of conditional generation.
}
   \label{fig:discriminator}
\end{figure}

We formulate the finetuning as an adversarial training process, where our video shader acts as the generator and we introduce a powerful critic network as the discriminator. Specifically, we build our discriminator with the frozen backbone of the pretrained WAN video diffusion model. We argue that this large model, trained on vast amounts of video data, already possesses a strong implicit prior of natural video statistics, making it an ideal critic. A small, trainable classification head is added to this backbone. To make the discriminator conditional, we integrate the reference image $I_{ref}$ into the training process  by inputing both feature maps extracted from real/fake videos and the reference image into the classification head. The architecture of our discriminator is shown in Fig.~\ref{fig:discriminator}. 

For the training objective, we employ a relativistic loss~\cite{jolicoeur2018relativistic} for stabler training. The training objectives for the discriminator $D$ (with parameters $\psi$) and the generator $G$ (with parameters $\theta$) are defined as:
\begin{align}
    \mathcal{L}_{D}(\psi) = & -\mathbb{E} \big[ \log ( \text{sigmoid}(D(V_{gt}, \cdot) - D(\hat{V}, \cdot))) \big] \notag \\
    & \qquad + \lambda_{reg} \mathcal{L}_{reg}, \label{eq:loss_d} \\
    \mathcal{L}_{G}(\theta) = & -\mathbb{E} \big[ \log ( \text{sigmoid}(D(\hat{V}, \cdot) - D(V_{gt}, \cdot))) \big], \label{eq:loss_g}
\end{align}
where $\cdot$ denotes the conditioning reference image $I_{ref}$ for brevity.
We also adopt the approximate R1, R2 regularization~\cite{huang2025self, lin2025autoregressive, lin2025diffusion} to stabilize training. 

\subsection{Progressive Training Strategy}
\label{sec:train_strategy}
To facilitate stable end-to-end training and enhance temporal consistency, we adopt a two-stage progressive training strategy.

\noindent \textbf{Stage 1: End-to-End Alignment.} We first initialize the reconstruction model through a brief warm-up phase and then train the entire framework, i.e., the reconstruction model and the video shader end-to-end on short video clips (e.g., 9 frames). This stage focuses on aligning the 3D-aware feature with the autoregressive video diffusion model. The total training objective is:
\begin{equation}
\mathcal{L}_{stage1} = \mathcal{L}_{recon} + \mathcal{L}_{pair} + \mathcal{L}_{adv}.
\end{equation}

\noindent \textbf{Stage 2: Temporal Extension.} After the initial alignment, we freeze the parameters of the 3D reconstruction model and continue training the video shader on longer sequences. This stage enhances the model's ability to maintain temporal consistency over extended periods. The training objective in this stage is:
\begin{equation}
\mathcal{L}_{stage2} = \mathcal{L}_{pair} + \mathcal{L}_{adv}.
\end{equation}
During finetuning, we employ the same self-rollout strategy as \mbox{Self-forcing}~\cite{huang2025self} to prevent train-test mismatch. To better preserve the rich knowledge of the pretrained model and enhance training efficiency, only the input layer for the 3D-aware features and  the self-attention blocks in the video diffusion model are finetuned, while other weights are freezed.

\begin{figure}[t]
  \centering
  \includegraphics[width=0.48\textwidth]{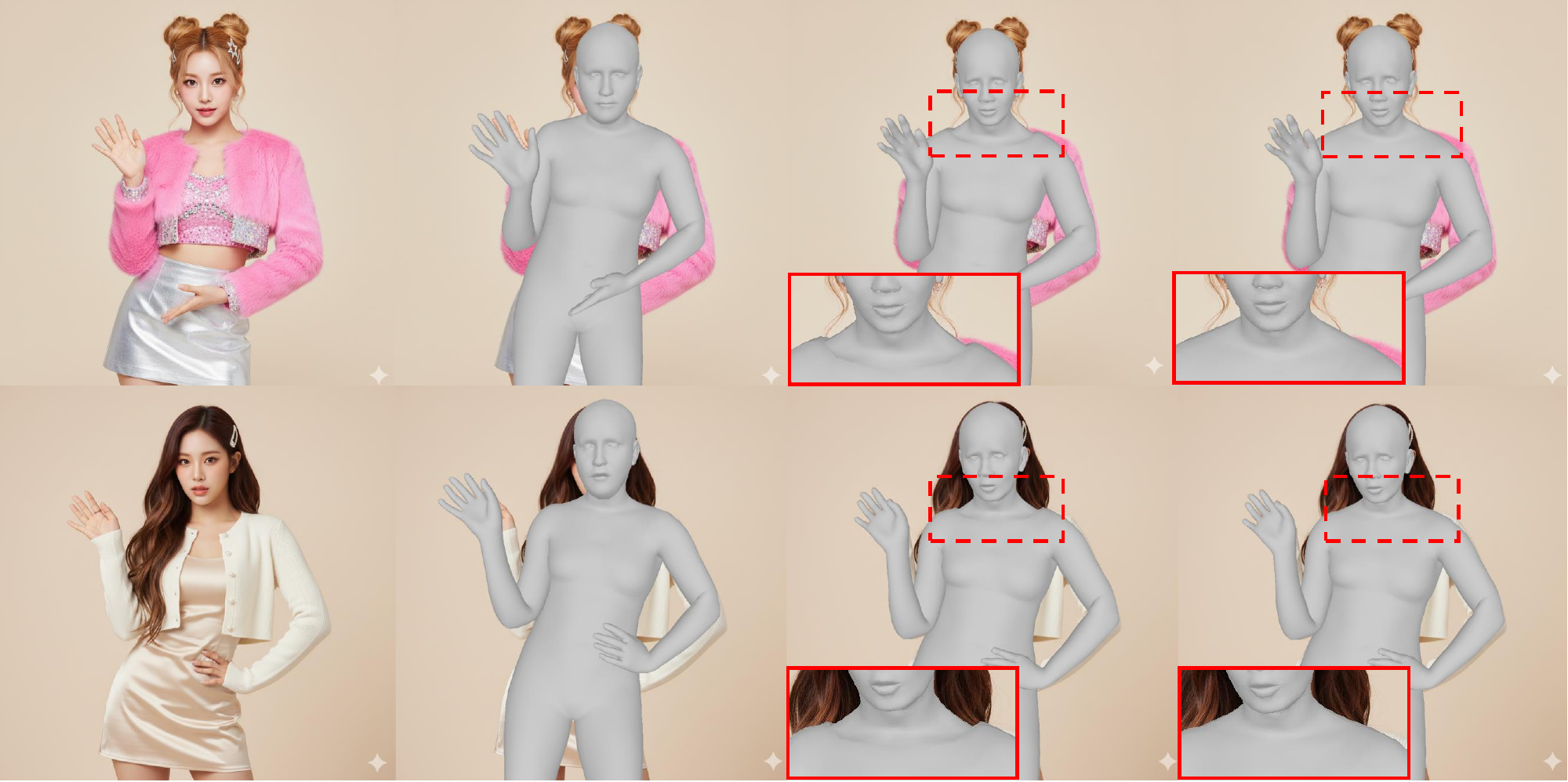}
    \makebox[0.11\textwidth]{\footnotesize Input Image}
    \makebox[0.11\textwidth]{\footnotesize Initialization}
    \makebox[0.11\textwidth]{\footnotesize w/o Remove}
    \makebox[0.11\textwidth]{\footnotesize w/ Remove}
  \caption{\textbf{The effects of our neck remove strategy}. This simple yet effective strategy eliminates neck artifacts, yielding a coherent, natural-looking full-body template for training. The \fcolorbox{red}{white}{red boxes} highlight the visual difference.}

\vspace{-1em} 

\label{fig:neck}
\end{figure}

\subsection{Dataset Curation}
\label{sec:dataset}

We construct our training dataset from Speaker-5M~\cite{zhang2025speakervid}, a large-scale dataset of high-quality human portrait videos. Our data processing pipeline involves several key stages to ensure high-quality annotations suitable for 3D-aware avatar generation.

\noindent \textbf{Initial Filtering.}
The pipeline begins by discarding static clips, identified by analyzing the temporal variance of 2D keypoint locations. As our focus is on upper-body avatars, we further select clips where the face and hands are clearly visible, retaining only those with corresponding keypoint confidence scores above a predefined threshold.

\noindent \textbf{Parameter Estimation and Refinement.}
We employ Multi-HMR~\cite{multi-hmr2024} and GAGATracker~\cite{chu2024gagavatar} to obtain initial estimates for SMPL-X and FLAME parameters, respectively. However, direct feed-forward estimation for in-the-wild, upper-body videos is notoriously challenging, often yielding inaccurate poses (particularly for hands), shapes, and expressions. To refine these estimates, we adopt a per-video optimization strategy similar to recent works~\cite{moon2024exavatar, GUAVA}. This involves minimizing the reprojection error between the 3D body/facial keypoints and their corresponding 2D detections. Following this optimization, a second filtering pass is performed to discard any clips where the final reprojection error remains above a specified threshold, ensuring a high-quality ground-truth dataset.

\noindent \textbf{Seamless Head-Body Fusion.}
To achieve enhanced facial expressiveness, we replace the head of the SMPL-X model with the more detailed FLAME model~\cite{GUAVA}. A naive replacement, however, often results in a noticeable discontinuity artifact at the neck seam due to geometric inconsistencies. To resolve this, we propose a refined fusion technique. Instead of swapping the entire head region, we selectively graft only the facial region of FLAME onto the SMPL-X model, while retaining the original SMPL-X neck.

Specifically, we identify the neck vertices of the FLAME model by leveraging its skinning weights, $W_{flame} \in \mathbb{R}^{N_{flame} \times J_{flame}}$. A vertex $v_i$ is classified as part of the neck if its maximally influencing joint is the neck joint:
\begin{equation}
    \underset{j}{\text{argmax}} \, (W_{flame}[i, j]) = id_{neck}
\end{equation}
By excluding these identified neck vertices from the FLAME model during the fusion, we create a seamless transition to the original SMPL-X neck. As shown in Fig.~\ref{fig:neck}, this simple yet effective strategy successfully eliminates the neck artifact, ensuring a coherent and natural-looking full-body template for our training.

\section{Experiments}
\label{sec:experiment}

\begin{figure*}[t]
  \centering
  \includegraphics[width=0.96\textwidth]{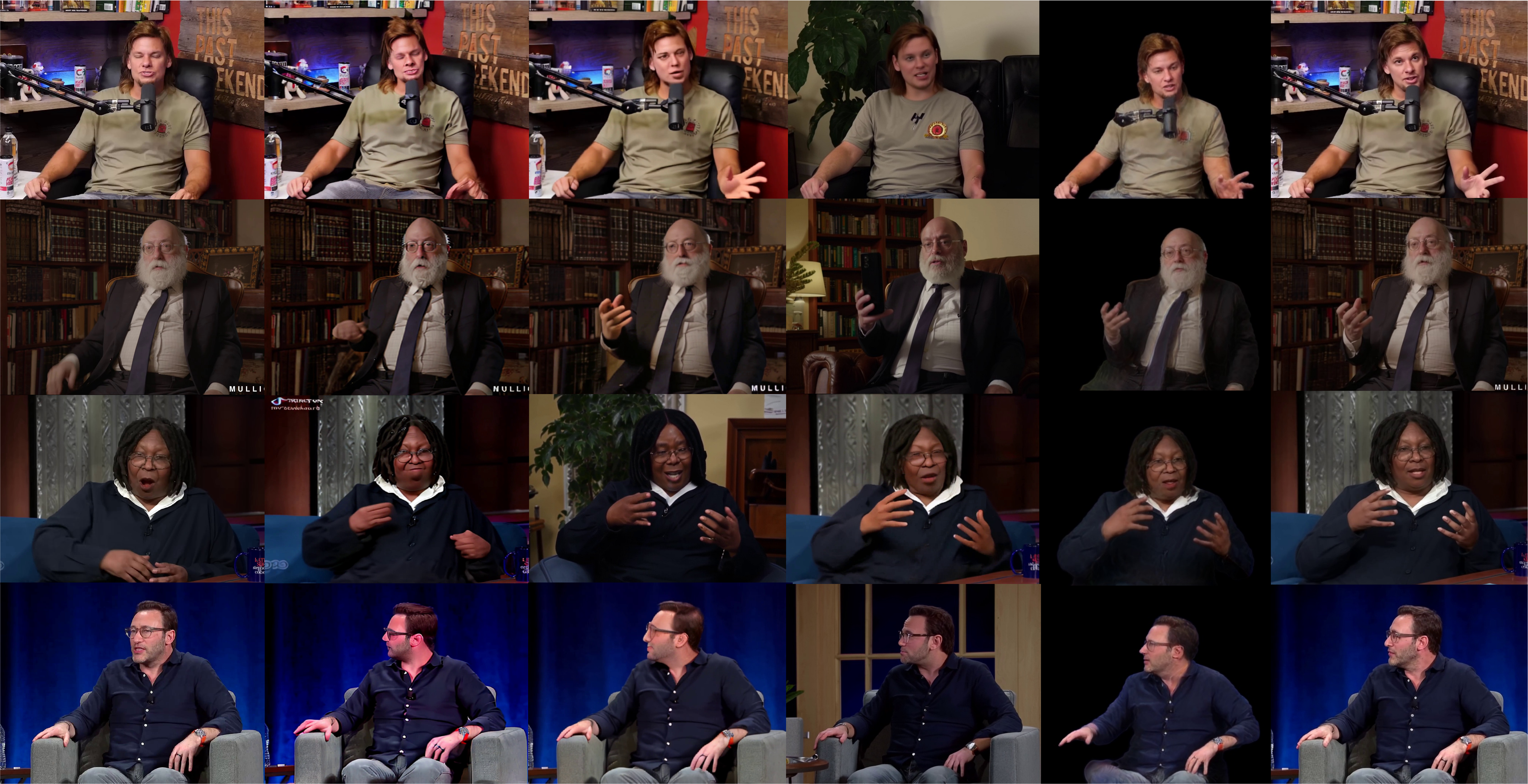}
    \makebox[0.13\textwidth]{\footnotesize Input Image}
    \makebox[0.19\textwidth]{\footnotesize Champ~\cite{zhu2024champ}}
    \makebox[0.14\textwidth]{\footnotesize MimiMotion~\cite{zhang2024mimicmotion}}
    \makebox[0.18\textwidth]{\footnotesize VACE~\cite{vace}}
    \makebox[0.11\textwidth]{\footnotesize GUAVA~\cite{GUAVA}}
    \makebox[0.13\textwidth]{\footnotesize Ours}

  \caption{\textbf{Qualitative comparisons on self-reenactment}. Compared with existing approaches, our method generates more accurate and clearer results.}
   \label{fig:self_reenact}
\end{figure*}

\begin{figure*}[t]
  \centering
  \includegraphics[width=0.96\textwidth]{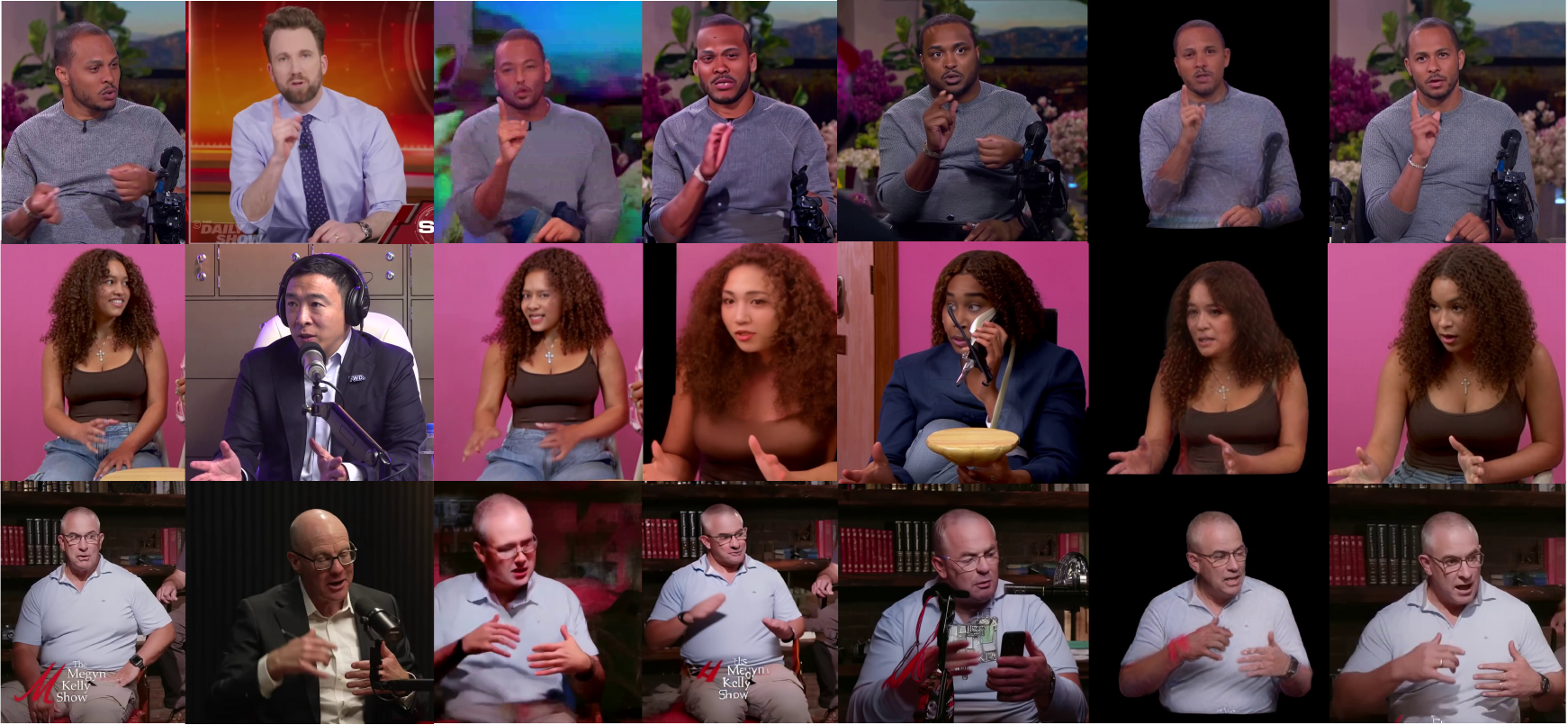}
    \makebox[0.12\textwidth]{\footnotesize Input Image}
    \makebox[0.13\textwidth]{\footnotesize Reference}
    \makebox[0.13\textwidth]{\footnotesize Champ~\cite{zhu2024champ}}
    \makebox[0.13\textwidth]{\footnotesize MimiMotion~\cite{zhang2024mimicmotion}}
    \makebox[0.13\textwidth]{\footnotesize VACE~\cite{vace}}
    \makebox[0.13\textwidth]{\footnotesize GUAVA~\cite{GUAVA}}
    \makebox[0.13\textwidth]{\footnotesize Ours}

  \caption{\textbf{Qualitative results on cross-reenactment against existing methods}. Our approach outperforms alternatives in maintaining identity consistency across a range of poses while more faithfully capturing facial expressions and fine-grained details.}
   \label{fig:cross_reenact}
\end{figure*}

\begin{figure}[t]
  \centering
  \includegraphics[width=0.48\textwidth]{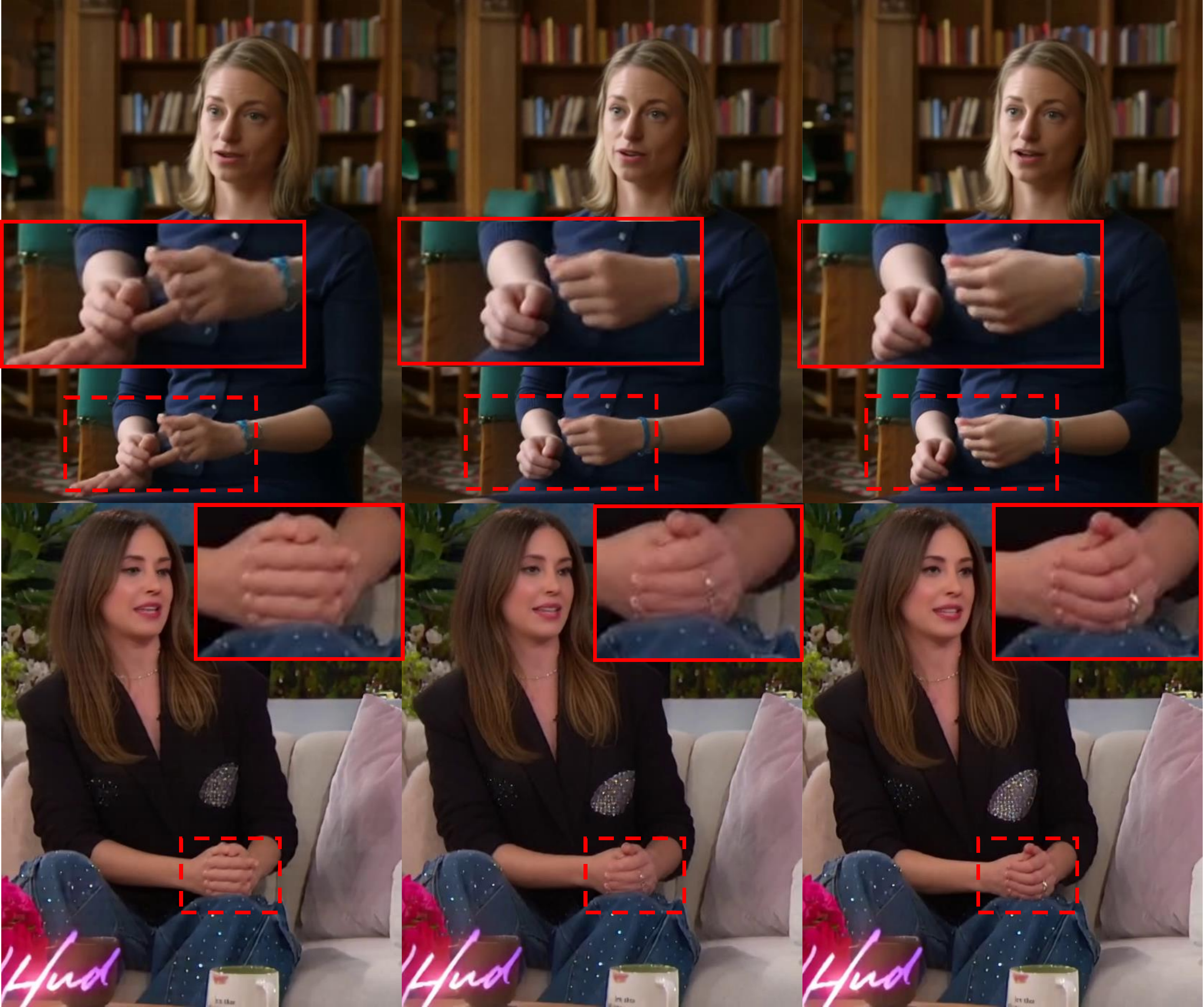}
    \makebox[0.15\textwidth]{\footnotesize (a) Keypoints }
    \makebox[0.15\textwidth]{\footnotesize (b) 3D-aware RGB }
    \makebox[0.15\textwidth]{\footnotesize (c) 3D-aware Features}

  \caption{\textbf{The ablation study for different condition inputs discussed in Sec.~\ref{sec:dc}}. The \fcolorbox{red}{white}{red boxes} highlight the main differences between the conditions.}
   \label{fig:ablation_cond}
   \vspace{-1em}
\end{figure}

\begin{figure}[t]
  \centering
  \includegraphics[width=\linewidth]{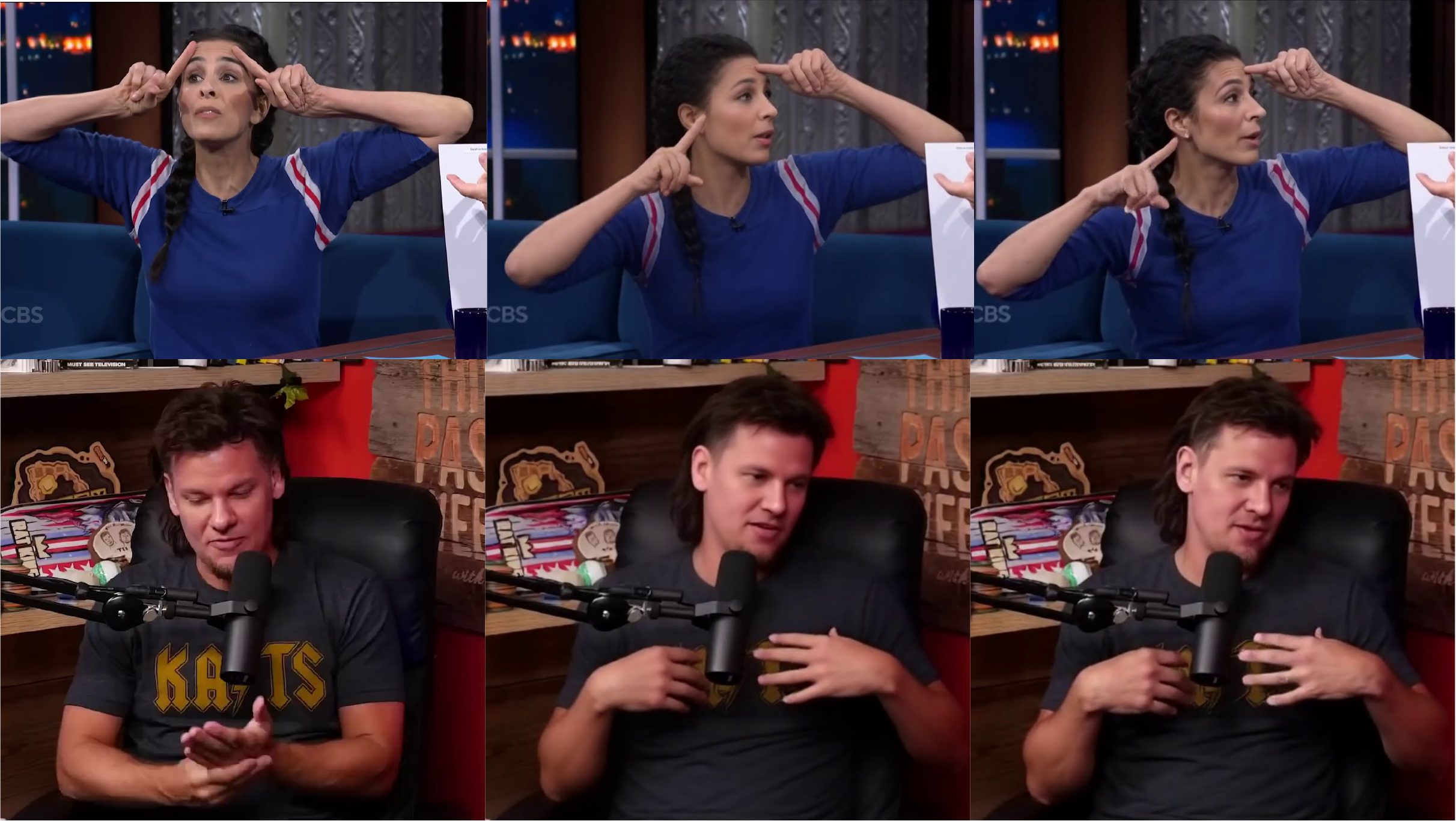}
    \makebox[0.15\textwidth]{\footnotesize Input Image}
    \makebox[0.15\textwidth]{\footnotesize w/o ADP}
    \makebox[0.15\textwidth]{\footnotesize w/ ADP}

  \caption{\textbf{The effects of Adversarial Distribution Preservation loss}. The figure demonstrates the Adversarial Distribution Preservation loss effectively improves visual quality, reducing blur and artifacts, especially in high-frequency regions. Please zoom in for better view.}
   \label{fig:ablation_adp}
\end{figure}


\subsection{Experimental Setup}

\noindent\textbf{Implementation Details.}
Using the curation procedure described in Section~\ref{sec:dataset}, we construct a dataset of 46,000 video clips, each containing approximately 150 frames on average (corresponding to a typical duration of 3–10 seconds). From this collection, we randomly select 100 subject IDs, each contributing one video clip, as the test set, with the remaining clips forming the training set. Our ViSA model is initialized from Self-forcing 1.3B model~\cite{huang2025self}  and trained on this training set using the Adam optimizer ($\beta_1 = 0.9$, $\beta_2 = 0.999$) with a learning rate of $5 \times 10^{-5}$. Stage one is trained on 32 NVIDIA H20 GPUs with a total batch size of 64 for 4,000 steps. Stage two is then trained for 6,000 steps using the same batch size and number of GPUs. The entire training process takes approximately five days.


\noindent\textbf{Metrics.}
\noindent\textit{Self-reenactment.}  
We use the first frame of each video as the reference image and animate it using the full motion sequence from the same video. To evaluate image quality, we compute PSNR, SSIM, and LPIPS between all generated frames and their corresponding ground-truth frames in the original video. Additionally, we measure Identity Preservation Score (IPS-self) by computing the cosine distance between ArcFace features extracted from the face regions of the generated and their corresponding original frames.
\noindent\textit{Cross-reenactment.}  
Since no ground-truth output exists when animating a source identity with motion from a different subject, we assess identity consistency using the Identity Preservation Score (IPS-cross), defined as the average cosine similarity between ArcFace features of the reference image and all frames in the generated video.



\begin{table}[t]
\centering
\small
\caption{Quantitative comparison of avatar generation methods.}
\label{tab:quantitative}
\resizebox{\columnwidth}{!}{%
    \begin{tabular}{lcccccc}
    \toprule
    Method & PSNR $\uparrow$ & SSIM $\uparrow$ & LPIPS $\downarrow$ & IPS-self $\downarrow$ & IPS-cross $\downarrow$ & Speed (fps) $\uparrow$ \\
    \midrule
    VACE~\cite{jiang2025vace} & 14.83 & 0.78 & 0.110 & 0.053 & 0.066 & 0.28 \\
    Champ~\cite{zhu2024champ} & 15.57 & 0.79 & 0.106 & 0.049 & 0.074 & 1.90 \\
    MimicMotion~\cite{zhang2024mimicmotion} & 17.91 & 0.83 & 0.087 & 0.045 & 0.065 & 0.72 \\
    GUAVA~\cite{GUAVA}      & 18.59 & 0.86 &  0.072 & 0.040 & 0.061 & \textbf{89.12} \\
    ViSA (Ours)   & \textbf{22.09} & \textbf{0.87} & \textbf{0.043} & \textbf{0.037} & \textbf{0.060} & 15.15  \\
    \bottomrule
    \end{tabular}%
}
\end{table}



\subsection{Comparison with State-of-the-art Methods}

We compare our ViSA with state-of-the-art \textit{2D-based methods}, including MimicMotion~\cite{zhang2024mimicmotion}, Champ~\cite{zhu2024champ}, and VACE~\cite{jiang2025vace}, and \textit{3D-based methods}, i.e. GUAVA~\cite{GUAVA}.

As shown in Fig.~\ref{fig:self_reenact} and Fig.~\ref{fig:cross_reenact}, video-based methods often suffer from identity drift: the generated avatar fails to maintain consistent appearance with the input reference image, and its identity gradually shifts over time. Additionally, pose details are frequently unstable. For instance, hands appear blurry or adopt anatomically implausible configurations. Backgrounds are also poorly handled; VACE tends to replace them with arbitrary scenes, while Champ hallucinates unrealistic elements such as a floating microphone. These issues are reflected in Table~\ref{tab:quantitative}, where these methods exhibit inferior PSNR, SSIM, and LPIPS scores, as well as higher IPS-self and IPS-cross distances indicating poor identity preservation. Moreover, their reliance on multi-step diffusion models incurs high computational cost, requiring several minutes to generate a short $10\sim20$ seconds clip.
In contrast, the 3D-based method GUAVA leverages efficient 3D Gaussian Splatting (3DGS) rendering to achieve real-time synthesis. By attaching Gaussians to an SMPL-X template, it ensures motion controllability and stable identity. However, this rigidity limits expressiveness: motions appear stiff, fine details are lost, and backgrounds cannot be synthesized.

Our approach bridges the gap between these paradigms. By integrating 3D-aware avatar features with a lightweight, few-step autoregressive video diffusion model, ViSA achieves both high fidelity and real-time performance. The 3D prior ensures identity consistency, pose accuracy, and coherent background integration, while the generative renderer enriches details and produces fluid motion. This synergy yields the best quantitative results in Table~\ref{tab:quantitative} and enables faster generation at 15 FPS on a single A100 GPU.




\subsection{Ablation Study} 
\noindent\textbf{Different Condition Inputs.} \Fref{fig:ablation_cond} shows the ablation study for different guidance inputs. For human keypoint inputs, the skeletal representation provides only coarse positional information and lacks pixel-level guidance for image synthesis. As a result, it produces severe artifacts in high-frequency regions, such as the fingers. In contrast, our model conditions on pixel-aligned guidance generated from a 3D reconstruction model, yielding more accurate and consistent animation results. Also, \fref{fig:ablation_cond}~(b)–(c) shows that the model conditioned on 3D-aware features can capture more precise semantic information and produces more natural, photorealistic animations than that conditioned on 3D-aware RGB. The component-wise speed analysis in Table~\ref{tab:speed} further compares the efficiency of feature-based condition injection versus traditional rgb-based injection. Our feature-based conditioning bypasses the VAE encoder, resulting in a 34\% acceleration over the RGB-conditioned pipeline.

\begin{table}[t]
\centering
\small
\caption{Component-wise speed analysis (ms per frame) of generation speed for different condition framework. The speed is tested on single A100 GPU.}
\label{tab:speed}
\setlength{\tabcolsep}{3.5pt} 
\begin{tabular}{lccccc}
\toprule
Condition & GS Render & Encode & Denoise & VAE & All \\
\midrule
Feature Condition & 2 & \textbf{-} & 62 & 2 & \textbf{66} \\
RGB Condition & 1 & 36 & 62 & 2 & 101 \\
\bottomrule
\end{tabular}
\vspace{-1em}
\end{table}

\noindent\textbf{The Effects of Adversarial Distribution Preservation} \Fref{fig:ablation_adp} presents an ablation study of our Adversarial Distribution Preservation loss~(ADP). The results show that ADP effectively improves overall visualization quality compared to training without it, particularly in high-frequency regions, and yields more natural renderings.

\section{Conclusion}
\label{sec:conclusion}

We presented ViSA, a real-time video shading framework for photorealistic 3D avatar generation that effectively bridges geometric reconstruction and generative video synthesis. By leveraging a 3D-aware reconstruction module to provide structural and appearance priors, our method guides a distilled autoregressive video diffusion model to render high-fidelity, temporally coherent avatars with rich details and natural motion. 
Experiments demonstrate that ViSA outperforms state-of-the-art approaches in both visual quality and inference speed. 



{
    \small
    \bibliographystyle{ieeenat_fullname}
    \bibliography{main}
}



\end{document}